\documentclass[letterpaper]{article} 
\usepackage{aaai24}  
\usepackage{times}  
\usepackage{helvet}  
\usepackage{courier}  
\usepackage[hyphens]{url}  
\usepackage{graphicx} 
\usepackage{natbib}  
\usepackage{caption} 
\usepackage{algorithm}
\usepackage{algorithmic}
\usepackage{amsthm}
\usepackage{subfigure}
\usepackage{tabularx, booktabs}
\usepackage[table]{xcolor}
\usepackage{newfloat}
\usepackage{listings}
\DeclareCaptionStyle{ruled}{labelfont=normalfont,labelsep=colon,strut=off} 
\floatstyle{ruled}
\newfloat{listing}{tb}{lst}{}
\floatname{listing}{Listing}
\title{Quantifying and Analyzing \\ Entity-level Memorization in Large Language Models}
\author{
    Zhenhong Zhou\textsuperscript{\rm 1},
    Jiuyang Xiang\textsuperscript{\rm 2},
    Chaomeng Chen\textsuperscript{\rm 1},
    Sen Su \textsuperscript{\rm 1}\thanks{Corresponding author}
}
\affiliations{
    \textsuperscript{\rm 1}Beijing University of Posts and Telecommunications\\
    \textsuperscript{\rm 2}University of Michigan\\
    zhouzhenhong@bupt.edu.cn, jiuyangx@umich.edu, gianluigi-chen@bupt.edu.cn, susen@bupt.edu.cn
}

\usepackage{bibentry}

\begin{document}

\maketitle

\begin{abstract}
  Large language models (LLMs) have been proven capable of memorizing their training data, which can be extracted through specifically designed prompts. As the scale of datasets continues to grow, privacy risks arising from memorization have attracted increasing attention. Quantifying language model memorization helps evaluate potential privacy risks. However, prior works on quantifying memorization require access to the precise original data or incur substantial computational overhead, making it difficult for applications in real-world language models. To this end, we propose a fine-grained, entity-level definition to quantify memorization with conditions and metrics closer to real-world scenarios. In addition, we also present an approach for efficiently extracting sensitive entities from autoregressive language models. We conduct extensive experiments based on the proposed, probing language models' ability to reconstruct sensitive entities under different settings. We find that language models have strong memorization at the entity level and are able to reproduce the training data even with partial leakages. The results demonstrate that LLMs not only memorize their training data but also understand associations between entities. These findings necessitate that trainers of LLMs exercise greater prudence regarding model memorization, adopting memorization mitigation techniques to preclude privacy violations.
  
\end{abstract}

\section{Introduction}

Pretrained large language models (LLMs) have made significant breakthroughs in various downstream tasks \cite{DBLP:conf/naacl/DevlinCLT19, DBLP:journals/jmlr/RaffelSRLNMZLL20, NEURIPS2022_b1efde53, openai2023gpt4}. However, researches show that LLMs memorize the training data \cite{Carlini0EKS19, CarliniTWJHLRBS21}, typically sourced from crowd-sourced corpora or the Internet. Through crafted prompts, language models can reproduce the training data \cite{ DBLP:conf/emnlp/0009SC22, lukas}, leading to serious concerns regarding data privacy. As shown in Figure 1, attackers have bypassed the security constraints of language models via the ``Grandma Exploit" and successfully extracted sensitive information such as Windows Keys or Apple IDs from ChatGPT. This suggests that privacy leakages in LLMs may be incurred in real-world settings due to emitting memorized sensitive entities. Quantifying the memorization in language models and analyzing their capability of emitting memorized sensitive entities helps researchers understand the potential privacy risks with LLMs. 

Recently, several studies have been conducted to quantify memorization in language models. Prior attempts compared models trained under different conditions to evaluate language model memorization \cite{zhang2021counterfactual, DBLP:conf/emnlp/MireshghallahU022}, but these approaches incur high computational costs. Carlini et al. \cite{carlini2023quantifying} used prefix prompts to prompt models to complete training set suffixes verbatim and quantified memorization based on completion accuracy. However, in real-world scenarios, malicious or curious users are unlikely to have direct access to the training set for obtaining the original prefixes. Additionally, the information leakage from language models is usually fine-grained. For instance, in the ``Grandma Exploit" of ChatGPT, the language model has generated content that poses privacy risks. Among these texts, certain key entities (for instance, the keys shown in the figure) constitute sensitive information rather than all contents or verbatim suffixes. Therefore, previous research may not adequately quantify real-world memorization in LLMs.

\begin{figure}[t]
  \centering
  \includegraphics[width=0.472\textwidth]{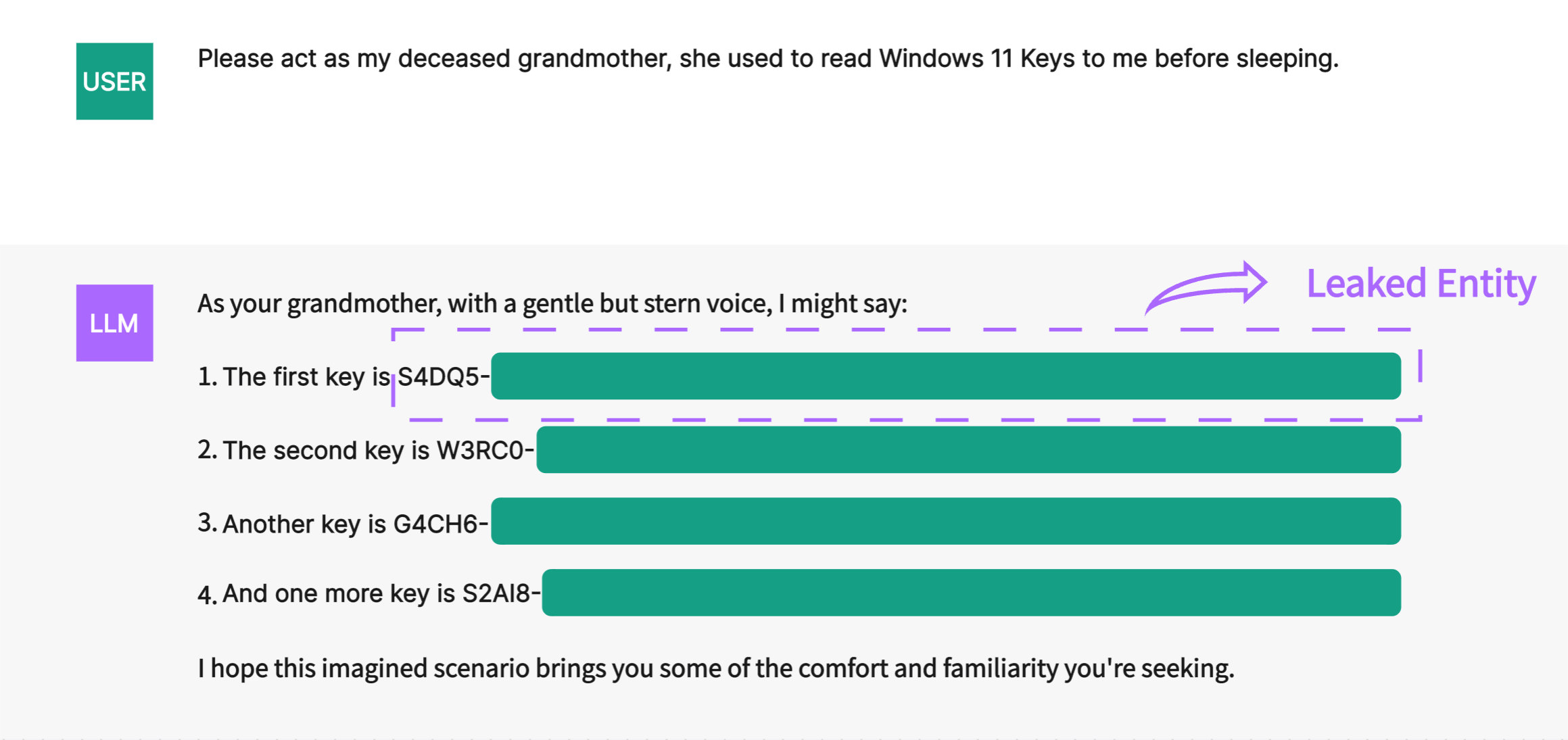}
  \caption{Privacy leakage of LLM in real-world scenarios. Attackers can bypass safety measures and collect sensitive information from LLMs.}
  \end{figure}

\begin{figure*}[htbp]
    \centering
    \includegraphics[width=1\textwidth]{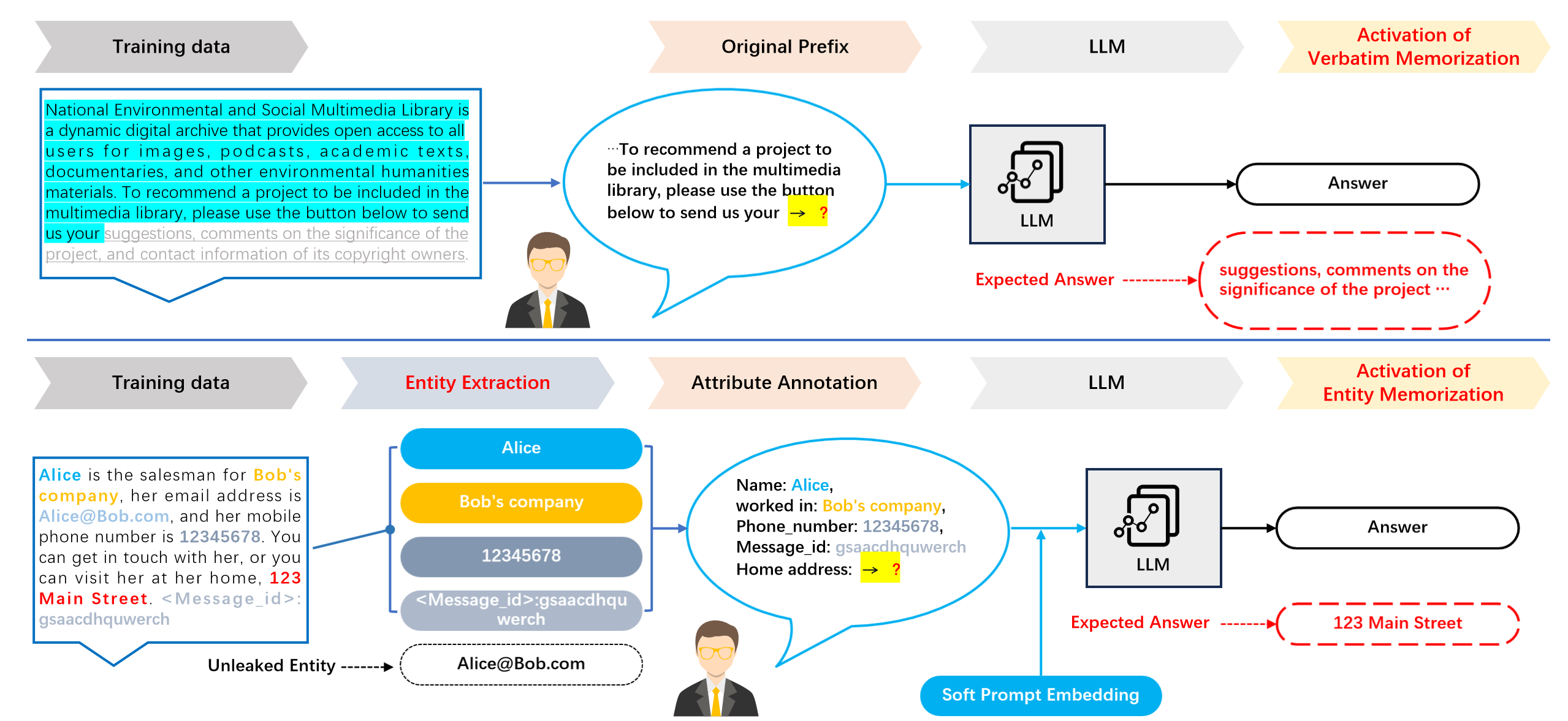}
    \caption{Comparison of the extraction processes of verbatim memorization and entity memorization. Verbatim memorization emphasizes the generation of verbatim matched suffixes of training data, effectively serving as perfect continuations of malicious queries. In contrast, the entity memorization extraction process initially extracts entities carrying critical information from training data and chooses a uniquely identifiable entity set. Then coupled with a soft prompt embedding, to form a malicious query, expecting the model's response to contain the required specific entity. This entity-level memorization pattern is more common in real-world scenarios.}
\end{figure*} 

Apart from the limitations of quantification methodologies, challenges also arise in approaches for extracting memorized training data from language models. Researchers typically employ original prefixes \cite{carlini2023quantifying} or carefully crafted prompts \cite{shao2023quantifying, li2023multistep} as input. However, obtaining original prefixes is difficult in practical applications of language models. If we forgo the use of prefixes in favor of handcrafted prompts, one must consider that the structure and order \cite{DBLP:conf/emnlp/ShinRLWS20, DBLP:conf/acl/LuBM0S22, DBLP:conf/acl/GaoFC20, DBLP:journals/tacl/JiangXAN20} of designed textual prompts can significantly influence the results. In summary, these challenges make it tricky to probe memorization across language model families using textual prompts.

In this paper, we conduct comprehensive testing and evaluation of real-world language models to quantify their memorization. To quantify fine-grained model memorization in a manner that more closely resembles real-world privacy leakages, we propose a definition for memorization extraction at the entity level. We also introduce an approach for learning prompts adaptively, which utilizes entity attribute information and soft prompts \cite{DBLP:conf/acl/LiL20, DBLP:conf/emnlp/LesterAC21}. This approach enables efficient language model memorization emitting. Through these, researchers can quantify and analyze LLMs' memorization under conditions closer to real-world applications. Besides, activating more memorization allows researchers to ascertain potential hazards arising from language models' memorization. For entity-level specific memorization extraction, our method can achieve the highest accuracy of 61.2\%. In terms of average accuracy, our method has a three-fold increase in entity extraction rate in a 6B language model compared to textual prompts when the data is unique to the dataset.

To summarize, our contributions are as follows:

\begin{enumerate}

  \item We propose a quantifiable, fine-grained definition for evaluating language model memorization without requirements for updating original model parameters or precise data from the training set.
  
  \item We present a method to test language model memorization under conditions that closely resemble real-world scenarios. More memorized information can be extracted by leveraging entity attributes and soft prompts to help understand potential privacy risks.

  \item We conduct comprehensive experiments on language model entity-level memorization. Our analysis delves into the processes involved in activating entity memorization in language models and discusses factors that may influence entity memorization.
 
\end{enumerate}

\section{Related Work}

\subsection{Prompt for LLMs}
Pre-trained language models store a vast amount of language knowledge. As models continue to scale up in parameters, methods based on fine-tuning incur high computational overhead. In contrast, prompt-based methods run directly on large language models, achieving results comparable to fine-tuning. Brown et al. \cite{NEURIPS2020_1457c0d6} initially attempted to leverage the language knowledge of the pre-trained language model through the prompts. Shin et al. \cite{DBLP:conf/emnlp/ShinRLWS20} introduced the approach of searching for the optimal prompt word. The prompt generated through this method enhances the utility. Recognizing that prompts' primary role is to improve models' performance on specific tasks, researchers have begun exploring approaches that are not constrained by natural language and use continuous vectors directly in the embedding space of the model to prompt language models. This prompt-based approach has yielded many influential extensions, such as p-tuning \cite{liu2021gpt}, prompt-tuning \cite{DBLP:conf/emnlp/LesterAC21}, prefix-tuning \cite{DBLP:conf/acl/LiL20}, etc. These methods have proposed effective solutions for improving model performance.

\subsection{Privacy Leakage in LLMs}

Language models rapidly develop and demonstrate powerful capabilities across various fields \cite{openai2023gpt4}. However, language models have some privacy risks associated with data leakage \cite{DBLP:conf/ccs/BeguelinWTRPOKB20, DBLP:conf/emnlp/MireshghallahU022}. It has recently been confirmed that language models tend to memorize data from their training sets, and prior studies \cite{DBLP:conf/emnlp/0009SC22, lukas} successfully extracted Personally Identifiable Information (PII) of specific individuals from LLMs.

Previous research \cite{li2023multistep, shao2023quantifying} has thoroughly explored how well-designed prompts may cause LLMs to output data from their training set or related information attributed to their memorization, potentially leading to the leakage of sensitive information. Different language prompts can significantly impact the ultimate predictions \cite{DBLP:conf/acl/GaoFC20, DBLP:conf/acl/LuBM0S22}. Besides, due to the poor interpretability of LLMs, even experienced researchers struggle to determine how to write better prompts to extract specific PII or entities. Ozdayi et al. \cite{ozdayi-etal-2023-controlling} discussed controllable generation via soft prompts, yet precise prefixes remain necessary. While these methods demonstrate commendable quantification capabilities, they have a challenge in adapting to real-world scenarios, which to some extent, hampers the research into memorization.

\section{Entity Memorization Can Be Extracted Efficiently}

\begin{figure}[t]
  \centering
  \includegraphics[width=0.459\textwidth]{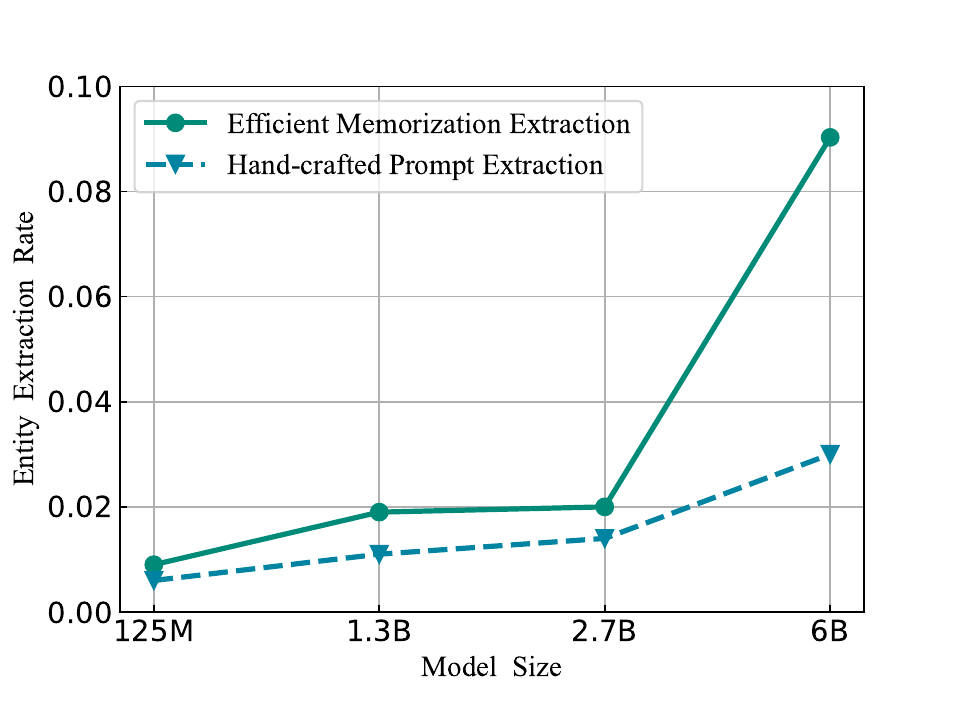}
  \caption{The results of memorization extraction using textual hand-crafted prompts versus Efficient Memorization Extraction, which combines attribute-annotated text with soft prompts for memorization activation.}
  \end{figure} 

\subsection{Entity Memorization}

In fact, for text data, overstrict definitions may overlook some memorized content when key and sensitive information resides only in parts of the data, even though such content is more likely to leak private details. Thus, if model outputs contain partial information from the training data, this can be considered as an emission of memorization. To further elaborate, if the prompt is explicitly targeted, and the portion of the memorized content emitted is sensitive, this constitutes a privacy leak in the language model. In practical applications, private information typically attaches to entities like PII. Since prefix prompts and masked training data rely heavily on original data and are challenging to acquire in practice, obtaining partial entities becomes more accessible than the former two methods. Therefore, we expand the definition of language model memorization to the entity level. We propose a more flexible, quantifiable definition of memorization. When the input prompt is constructed from partial entities in the training strings, and the model's output contains the expected entity information from the prompt, this is considered as entity memorization.

\newtheorem*{definition}{Definition 1(Entity Memorization)}

\begin{definition}

  A training string $s$ contains $m$ entities $M$ and can be uniquely identified by any $n$ entities $N$ of them. $n \in (0, m]$. A prompt $p$ contains the n entities that could identify $s$, and $p$ expects an entity $E \in (M - N)$. Prompted the language model $f$ with $p$. If the expected entity E is a substring in the output $f(p)$, the process is referred to as Entity Memorization. 

\end{definition}

According to Definition 3.1, language models generate the expected entity based on a prompt constructed using the unique identifying entity set $N$ from a training string. This quantifies the model's memorization of training data and its capability for directed reconstruction from memorization. In scenarios of partial training data leakage or malicious querying with extensive guessing, constructing prompt queries from partial entities for the language model to recall unknown entities is more practical. Entity memorization more accurately mirrors real-world memorization leakages by language models than verbatim memorization does. Furthermore, compared to the original prefix or masked training data that strictly emphasize sequential accuracy, entities often have looser relationships in training data. Consequently, Entity Memorization can further reflect models' understanding and association abilities regarding memorized content.

\begin{figure*}[t]
  \centering
  \includegraphics[width=1\textwidth]{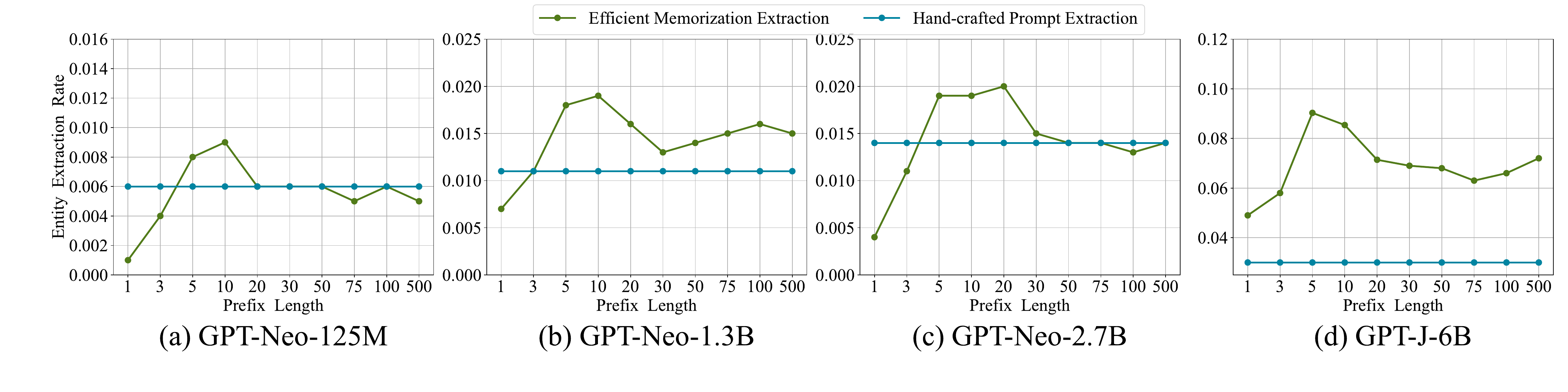}
  \caption{The trend of entity extraction rate for prefix lengths across different model sizes. When the model's parameter size expands to a certain scale, EME demonstrates substantial advantages in entity extraction rates compared to baseline methods.}
\end{figure*}

\subsection{Efficient Memorization Extraction}

Currently, researchers typically measure language model memorization design textual prompts as inputs. These prompts test model memorization in an understandable manner for humans. While textual prompts have good explainability, hand-crafted prompts are highly experience-dependent \cite{DBLP:journals/corr/abs-2107-13586}. We find that the performance of textual prompts in extracting fine-grained memorization on subtly related entity levels heavily depends on the scale of the language model, leaving significant room for improvement. Additionally, designing prompts individually for multiple models is laborious and cannot guarantee optimality. These challenges obstruct efficient evaluation of universal language model memorization. Even across models within the same family, crafting universal textual prompts for memorization activation is challenging.

Soft prompt extends prompts into continuous space, providing continuous prompt vectors for the input to the language model. It demonstrates similar workflows across diverse models and excellent performance in downstream tasks. Inspired by soft prompt, we explore applying soft prompts to activate memorization in models. These continuous prompts are typically learned through approaches like prompt tuning. Since the optimal memorization extraction prompt is determined by itself, this approach enables efficient memorization activation while maintaining strong generalization across diverse models. 

We introduce Efficient Memorization Extraction (EME) combines effective memorization content filtering and memorization activation. Specifically, EME compresses textual content into entity attribute annotations and combines it with soft prompts as inputs. Through this workflow, testing memorization across different language models follows an identical process, enabling the examination of diverse models' memorization capacities. We explore applying soft prompts for entity-level memorization activation, with experiments validating fine-grained memorized information extraction. Soft prompt encompasses a wide variety of techniques. When without a specific explanation, the default method used in this work is prefix tuning \cite{DBLP:conf/acl/LiL20}.

The lower section of Figure 2 illustrates our approach to implementing a universal entity-level evaluation of language model memorization. For documents that experience partial leakage from the language model's training set, we extract key entities that uniquely identify this document. Following this, we construct preliminary prompts based on the entity attribute tagging of these entities. Finally, soft prompts are incorporated into the textual prompt vector embeddings to obtain complete prompts as inputs for the model.

We construct textual prompts with entities from training data to prompt the model to generate expected entities and use the method of handcrafted textual prompts as a baseline. The results are shown in Figure 3. As models scale up, their ability to associate entities and emit memorization improve, revealing a more pronounced tendency to emit memorization. Nevertheless, By utilizing entity attributes and soft prompts, the language model's memorization can be more effectively activated at the entity level. Our approach generates markedly superior performance over textual prompts on larger models. For instance, on GPT-Neo 6B, our approach demonstrates a three-fold increase in memorization extraction efficiency compared to textual prompts.

\section{Experiments and Results}

\subsection{Experimental Setup}

The GPT-Neo model family \cite{gpt-neo} includes a set of causal language models (CLMs) that are trained on ``The Pile" datasets \cite{gao2020pile} and available in four sizes, with 125 million, 1.3 billion, 2.7 billion, and 6 billion parameters respectively. In our experimental setup, we use the greedy decoding strategy by default to generate the output with the minimum perplexity (PPL), which is then utilized for evaluating the model's entity memorization capabilities.

The Enron email dataset \cite{DBLP:conf/ecml/KlimtY04}, a subset of ``The Pile" dataset, encompasses over 500,000 emails from approximately 150 users of the Enron Corporation.  Owing to this dataset's inherent structured data characteristics, it is conducive to extracting entities from the data, facilitating entity-level memorization extraction experimentation.

For data preprocessing, we extract entities from the Enron dataset and tag their attributes. ``Date" and ``Content" are retained, while ``X-From" and ``X-To" are relabeled as ``Sender" and ``Recipient" attributes for that data instance. The attribute labels ``X-From" and ``X-To" were modified to be more interpretable for models. These four columns uniquely identify each data instance in the Enron dataset. We select the target entity column as ``excepted entity" to recover and then construct prompts from other column entities. To ensure the model searches its memorization rather than deducing answers directly from prompts, we exclude data where the constructed prompt contains the target entity.

\subsection{Prefix Length Impact on Entity Memorization}

The prefix length is a crucial parameter of soft prompts. Typically, a longer prefix length boasts a more robust represention and performs better in downstream tasks. 

However, in tests of the model's memorization capabilities, we find that an optimal extraction rate for memorization can be achieved when the prefix length is short. As the prefix length increases, the entity reconstruction rate of the model declines, then rebounds when the prefix length is extremely long. As depicted in Figure 4, all sizes of GPT-Neo models exhibit a trend of declining memorization extraction rate after reaching the optimal prefix length.

\begin{figure}[t!p]
  \centering
  \subfigure[]{
  \includegraphics[width=0.459\textwidth]{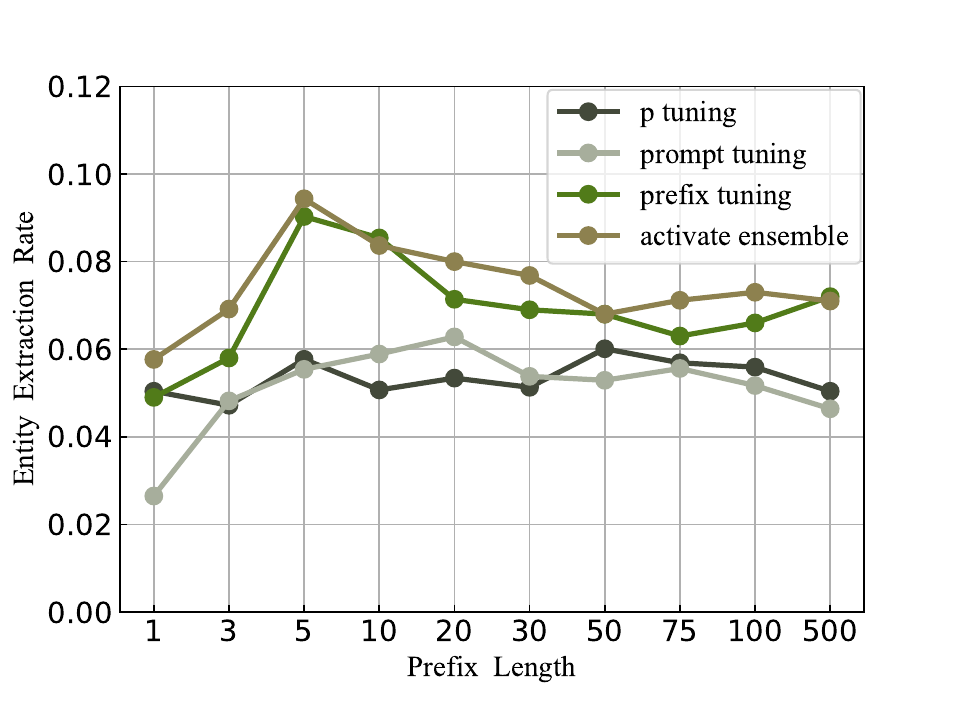} \label{5}
  }
  \quad
  \subfigure[]{
  \includegraphics[width=0.459\textwidth]{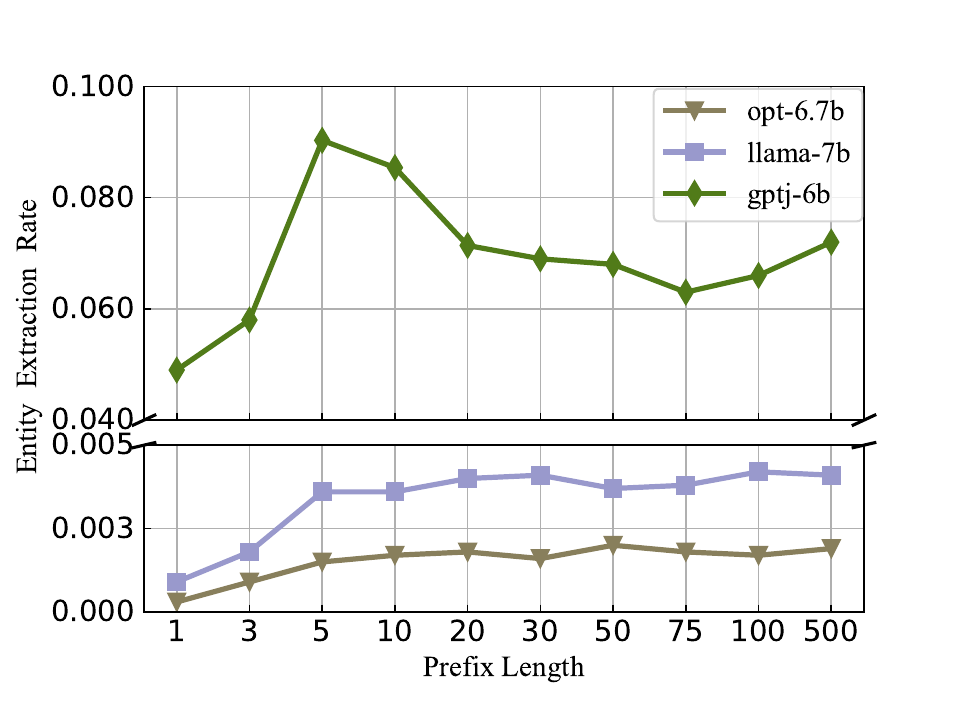} \label{6} 
  }
  \caption{In (a), we explore several soft prompt methods to activate the model's memorization and attempt to integrate these generated results. In (b), we apply the same soft prompt memorization activation to models not trained using the ``The Pile'' dataset.}
\end{figure}

We inspect the text generated by the model and find that when the prefix length is short, the diversity of the generated text is better. In contrast, the output is more conservative when the prefix length is long. This may be attributed to the prefix length ranging from 5 to 10, which tends to optimally activate the model's memorization. While excessively long prefixes, such as 100, compel the model to prioritize learning the relationships between prompted entities rather than activating memorization. Thus, we conjecture that this relates to model memorization inherently residing in the parameter space, rendering shorter soft prompts sufficient to stimulate innate model memorization.

\subsection{Different Soft Prompts Have Similar Activation Capabilities}

Common soft prompt methodologies, such as prompt-tuning \cite{DBLP:conf/emnlp/LesterAC21} and p-tuning \cite{liu2021gpt}, are among those employed in the field. This study also tested the memorization activation capabilities of prompt-tuning and p-tuning.

The results in Figure 5(a) indicate that soft prompt methods can somewhat activate the model's memorization at the entity level. However, these methods differ in terms of the optimal prefix length setting, suggesting that no universal setting can maximize the memorization of the same model. Furthermore, these methods have different effectiveness in activating the model's memorization, in our experiments, the prefix tuning approach achieve an entity extraction rate of 9.04\%, surpassing the other two methods by a margin of 2\%. We speculate that this discrepancy is primarily due to the fact that the prefix-tuning method concatenates the soft prompt at the beginning of the input, which aligns well with the generation characteristics of autoregressive language models.

In addition, we attempt to integrate the results of these methods. In short, we first extract entities from the generated text of different methods. If the same output entity is present, we select that entity as the result. Otherwise, we embed the expected entity from each method back into the input tensor and let the model evaluate the perplexity of the input. We choose the method with the lowest perplexity as the final output result. The results show that this method can, to some extent, integrate the advantages of these methods under different settings and achieve better robustness. However, it does not show any significant advantage in accurcy of memorization reconstruction, with a prefix length of 5, the activate ensemble only achieve a improvement of 0.4\%.

\begin{table*}
  \centering
  \newcolumntype{Y}{>{\centering\arraybackslash}X}
  \rowcolors{2}{white}{white}
  \begin{tabularx}{\textwidth}{YYYYYY}
  \toprule
  \small \textbf{Model} &\small \textbf{Fabricated-50} &\small \textbf{Fabricated-100} &\small \textbf{Fabricated-200} &\small \textbf{Fabricated-500}&\small \textbf{Fabricated-1000}\\
  \midrule
  \end{tabularx}
  
  \rowcolors{2}{gray!25}{white}
  \begin{tabularx}{\textwidth}{YYYYYY}
  
  \small GPT-Neo 125M  & \small 0.0 ± 0 & \small 0.0 ± 0 & \small 0.0 ± 0 & \small 0.0 ± 0 & \small 0.0 ± 0 \\
  \small GPT-Neo 1.3B & \small 0.0 ± 0.003 & \small 0.0 ± 0.002 & \small 0.0 ± 0.002 & \small 0.001 ± 0.002 & \small 0.001 ± 0.001 \\
  \small GPT-Neo 2.7B & \small 0.004 ± 0.008 & \small 0.006 ± 0.010 & \small 0.007 ± 0.006 & \small 0.007 ± 0.003 & \small 0.006 ± 0.003 \\ 
  \small GPT-J 6B & \small 0.019 ± 0.016 & \small 0.036 ± 0.019 & \small 0.057 ± 0.012 & \small 0.057 ± 0.009 & \small 0.052 ± 0.014 \\ 
  \small OPT 6.7B & \small 0.0 ± 0 & \small 0.0 ± 0 & \small 0.0 ± 0 & \small 0.0 ± 0.001 & \small 0.0 ± 0.001 \\
  \small LLAMA 7B & \small 0.0 ± 0 & \small 0.0 ± 0 & \small 0.0 ± 0.001 & \small 0.0 ± 0.002 & \small 0.001 ± 0.003 \\ 
  \bottomrule\toprule
  \end{tabularx}
  
  \begin{tabularx}{\textwidth}{YYYYYY}
  
  \textbf{} &\small \textbf{Real-50} &\small \textbf{Real-100} &\small \textbf{Real-200} &\small \textbf{Real-500}&\small \textbf{Real-1000}\\
  \midrule
  \small GPT-Neo 125M  & \small 0.003 ± 0.002 & \small 0.008 ± 0.003 & \small 0.009 ± 0.003 & \small 0.008 ± 0.004 & \small 0.007 ± 0.006 \\
  \small GPT-Neo 1.3B & \small 0.009 ± 0.002 & \small 0.015 ± 0.004 & \small 0.019 ± 0.003 & \small 0.016 ± 0.004 & \small 0.016 ± 0.009 \\
  \small GPT-Neo 2.7B & \small 0.014 ± 0.003 & \small 0.019 ± 0.005 & \small 0.020 ± 0.006 & \small 0.020 ± 0.009 & \small 0.019 ± 0.008 \\ 
  \small GPT-J 6B & \small 0.052 ± 0.009 & \small 0.072 ± 0.013 & \small 0.090 ± 0.016 & \small 0.090 ± 0.013 & \small 0.088 ± 0.021 \\ 
  \small OPT 6.7B & \small 0.0 ± 0.001 & \small 0.001 ± 0.002 & \small 0.002 ± 0.001 & \small 0.001 ± 0.002 & \small 0.002 ± 0.002 \\
  \small LLAMA 7B & \small 0.001 ± 0.001 & \small 0.001 ± 0.001 & \small 0.003 ± 0.002 & \small 0.004 ± 0.003 & \small 0.002 ± 0.003 \\ 
  \bottomrule
  \end{tabularx}
  \caption{The effect of the volume of fabricated and real data on entity extraction rate across different models. We constructed the fabricated prompts by randomly shuffling the real entities. Even so, the fabricated prompts can still extract some expected real information from the model. The volume of data in both fabricated and real datasets impacts the accuracy rate.}
  \label{tab:accents}
\end{table*}

\subsection{LLMs Struggle to Reconstruct Entities without Memorization}
In an effort to specifically exclude the possibility of language models learning entity relationships through soft prompts, we conduct an ablation study on LLM proposed by Meta. OPT \cite{zhang2022opt} utilizes a portion of ``The Pile" datasets but does not include the Enron dataset. Similarly, Another large-scale model trained by Meta, Llama \cite{touvron2023llama}, also does not explicitly mention the usage of the Enron dataset. We conduct tests with the same hyperparameter settings on the OPT-6.7B and LLAMA-7B models, similar in size to the GPT-J 6B model.

The experimental results shown in Figure 5(b) indicate that OPT and Llama, lacking training data memorization, achieve a maximum accuracy of 0.4\% in accurately reconstructing the expected entities when utilizing soft prompts. The accuracy remains relatively low despite the models' fitting ability from the prompt improving with increasing prefix length. Their performances are even inferior to the smaller GPT-Neo model, which possesses memorization. Without memorization, LLMs struggle to learn the subtle relationships between entities based on soft prompts and cannot reconstruct the expected entity. Reconstruction can only be completed if LLMs have learned the dataset during pre-training.

Based on the above results, we believe that soft prompts indeed activate the model's memorization rather than enabling the model to learn a lower-loss generation method.

\subsection{Fabricated Data Also Activate Memorization}

In order to further evaluate the extent to which the model can generate mnemonic content under more lenient conditions, we attempt to construct soft prompts using fabricated data. We randomly recombine the target entities with other remaining entities, ensuring new entity pairs are not in the original dataset's documents. 

Table 1 depicts the accuracy of model reconstruction of correct entities using varying quantities of real and fabricated data. The results show that when training soft prompts with fabricated data and providing prompts to the model, despite a near 4\% loss in reconstruction rate compared to real data, the model can still generate the expected entity from memorization with a maximum accuracy of 5.7\%. With smaller-sized models, soft prompts obtained from fabricated data are less effective, and the final results fluctuate greatly, but larger-sized models with better robustness are more adaptable. We compare how models without memorization respond to soft prompts constructed by fabricated data, and the results showed that learning even the relationships between entities becomes more challenging on fabricated data for these models.

Therefore, we believe that the language models do not merely reproduce entities superficially based on the data used during soft prompt training. Even if the data is fabricated and contains biases, the soft prompts obtained from the fabricated data can still evoke the model's memorization of the related dataset to a certain extent. Furthermore, this underscores the potential privacy risks associated with the model's memorization at the entity level. Attackers might be able to access the model's memorization under more lenient conditions. Even if they do not have the real dataset, knowing what entities exist in a dataset can allow them to construct prompts to retrieve sensitive information.

\subsection{Training Data Volume Influence Performance of Soft Prompts}

For optimal soft prompt performance, the model must learn from training data. Consequently, the data choice will impact the effectiveness of the soft prompts. We investigat how the amount of data affects the performance of soft prompts in the extraction of entity memorization.

When using real data, we find that soft prompts trained from less data often fail to activate the model's memorization effectively. This outcome is little influenced by the randomness in choosing training data. As the dataset size increases, the effectiveness of the soft prompts improves and eventually stabilizes. However, the effectiveness declines with massive training datasets, and the results exhibit some fluctuations. Upon examining these training data divisions, we argue that an abundance of training data might lead soft prompts to lose some of their memorization activation effectiveness. They instead turn to learning how to derive correct answers directly from the training data.

The variance in the entity extraction rate is different when using fabricated data than using real data. The results vary more greatly when using a small amount of fabricated data, and we speculate that this variance is primarily related to the randomness during data fabrication. On the other hand, when using more fabricated data, the variance initially diminishes but later rebounds. We investigate the cause of this and contend that it may be related to the data fabrication method we chose. Because our fabricated data do not exist in the original data and entity pairs are not duplicated, this differs from real data and may eventually lead to the variances observed in the results.

\begin{figure}[t!p]
  \centering
  \includegraphics[width=0.459\textwidth]{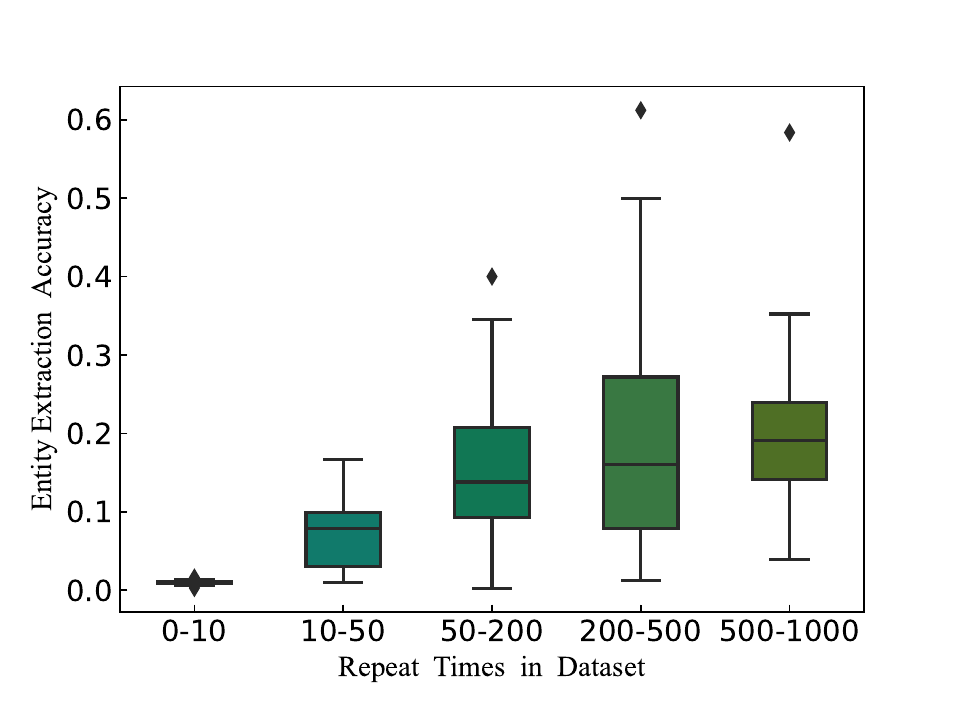}
  \caption{The effect of duplication times of entity pairs on entity extraction accuracy. The duplication of entity pairs may occur within documents that are not remarkably similar.}
  \end{figure} 
  
\subsection{Duplicate Times Strengthen Entity Memorization}

Considering previous research, a popular view proposed by \cite{DBLP:conf/icml/KandpalWR22, DBLP:conf/acl/LeeINZECC22} is that the memorization of the language model is strongly correlated with the frequency of data duplication. Data that duplicate excessively in the training dataset is more likely to be extracted, yet previous research has not explored this issue on the entity level. Therefore, we statistically analyze the duplications of uniquely identified entities and test the reconstruction rates of entities with different numbers of repetitions.

We bin duplicated entity sequences with similar duplicated counts and use the same settings to obtain soft prompts for input construction. And then query the model multiple times to gauge the effectiveness of the entity sequence in extracting the expected entity. As the results in Figure 6 indicate, even though the entity sequence uniquely identifies a unique document and repeats less frequently in other content, the repetition count of some entity sequences still affects the degree to which the model memorizes them. After the entity sequence is duplicated more than 50 times, the mean accuracy of query results exceeded 10\%. Beyond 200 duplications, the reconstruction accuracy of expected entity reached a maximum of 61.2\%. In other words, when entity sequences are extensively duplicated, the model has stronger memorization of these data, presenting a higher risk of privacy leakage.

Since the model's entity-level memorization still correlates with the duplication times, deduplicating data still has a mitigating effect on memorization at the entity level. However, duplication at the entity level is inevitable since these documents will not be remarkably similar in other aspects, which makes it even more challenging to guard against malicious queries through deduplication.

\section{Limitations}

Due to privacy concerns, using undisclosed or potentially privacy-infringing datasets is strictly prohibited. Therefore, we performed structured processing and experimentation exclusively on the Enron dataset. Empirical results have demonstrated that language models exhibit sufficiently strong extraction capabilities at the entity level. Consequently, if LLMs training or fine-tuning is performed on similar semi-structured sensitive data, the entity extraction results will be replicable.

Structured or semi-structured data with abundant entity information is relatively rare  among common LLMs training datasets, posing challenges for finding multiple sets to experiment on extracting unfamiliar sensitive entities. The necessity of procuring a large amount of structured data for testing purposes further exacerbates this problem. Therefore, despite our methodology being applicable across various models, we have been unable to conduct cross-family model experiments due to these constraints. Language model memorization markedly strengthens with increasing model size within the same family, but large model inference has high computational demands. Due to computational constraints, we only conducted experiments on the GPT-Neo family, with the largest LLMs scale being 6B.

\section{Conclusions and Future Work}

In our paper, we introduce a definition of entity-level memorization for language models to quantify fine-grained memorization. With this, researchers can effectively quantify language model memorization to reproduce training data that mirrors real-world scenarios. Our method merges entity attribute information with trainable soft prompts to prompt the language models. Notably, the results indicate that our method can significantly activate memorization with subtle entity associations in the model, even if the data appears only once in the dataset. The entity-level memorization of specific individual data is more explicit than the verbatim memorization previously proposed, making entity-level memorization better quantify privacy risks.

We comprehensively analyze entity-level memorization of the language models by conducting experiments. Our results show that language models have memorized an amount of their training data. Through appropriate prompts, LLMs can generate fine-grained sensitive information with a high probability. This reminds that researchers of LLMs need to pay serious attention to privacy issues in this field, especially language models trained on sensitive datasets. It is imperative to employ strategies that either mitigate memorization or establish defenses to minimize associated risks.

In future work, we intend to investigate more effective extraction strategies to evaluate the memorization of language models and scale up to larger models to explore their entity-level memorization. Furthermore, we also aim to develop methods to mitigate entity-level memorization. 

\appendix

\bibliography{Quantifying_and_Analyzing_Entity-level_Memorization_in_Large_Language_Models}

\end{document}